\documentclass[9pt,conference]{IEEEtran}
\IEEEoverridecommandlockouts
\usepackage{algorithm}
\usepackage{enumitem}
\setlist[itemize]{align=parleft,left=1pt..1em}
\usepackage{tikz}
\usetikzlibrary{arrows,automata,positioning,calc,arrows,arrows.meta}
\usepackage{wrapfig}

\PassOptionsToPackage{hyphens}{url}
\usepackage{hyperref}

\usepackage{algorithm}
\usepackage[noend]{algpseudocode}
\usepackage{cite}
\usepackage[utf8]{inputenc} 
\usepackage[T1]{fontenc}    
\usepackage{hyperref}       
\usepackage{url}            
\usepackage{booktabs}       
\usepackage{amsfonts}       
\usepackage{nicefrac}       
\usepackage{microtype}      
\usepackage{xcolor}         
\usepackage{listings}       
\usepackage{amsmath}

\usepackage{multirow}
\usepackage{tkz-kiviat} 
\usetikzlibrary{arrows}
\graphicspath{ {./figures/} }

\usepackage{pgfplots}
\pgfplotsset{compat=1.17}
\usepackage{caption}
\usepackage{subcaption}
\usepackage{indentfirst}
\usepackage{bm}

\newcommand{\tocite}[1]{\textcolor{red}{[TO CITE]}}

\def\BibTeX{{\rm B\kern-.05em{\sc i\kern-.025em b}\kern-.08em
    T\kern-.1667em\lower.7ex\hbox{E}\kern-.125emX}}
\begin{document}

\title{TD-Interpreter: Enhancing the Understanding of Timing Diagrams with Visual-Language Learning}


\author{
    Jie He$^{\dag}$, Vincent Theo Willem Kenbeek$^{\dag}$, Zhantao Yang$^{\S}$, Meixun Qu$^{\dag}$, Ezio Bartocci$^{\dag}$, Dejan Ničković$^{\ddag}$, Radu Grosu$^{\dag}$ \\
    \small $^{\dag}$Technische Universität Wien, Austria \\
    \small $^{\S}$Shanghai Jiao Tong University, China \\
    \small $^{\ddag}$AIT Austrian Institute of Technology, Austria 
}





\maketitle
\begin{abstract}

We introduce TD-Interpreter, a specialized ML tool that assists engineers in understanding complex timing diagrams (TDs), originating from a third party, during their design and verification process. TD-Interpreter is a visual question-answer environment which allows engineers to input a set of TDs and ask design and verification queries regarding these TDs. We implemented TD-Interpreter with multimodal learning by fine-tuning LLaVA, a lightweight 7B Multimodal Large Language Model (MLLM). To address limited training data availability, we developed a synthetic data generation workflow that aligns visual information with its textual interpretation. Our experimental evaluation demonstrates the usefulness of TD-Interpreter which outperformed untuned GPT-4o by a large margin on the evaluated benchmarks.
\end{abstract}

\newcommand\footnoteref[1]{\protected@xdef\@thefnmark{\ref{#1}}\@footnotemark}


\section{Introduction}
\label{sec:intro}

Digital design is a complex process involving the transformation of requirements into digital circuits by using hardware description languages. While High-Level Synthesis methods and advanced IP cores reduce the time and complexity of the design process, familiarity of engineers with the function of the various hardware components remains essential. Timing diagrams (TDs), as a visual requirements specification of the functionality and interaction between different signals, play a central role in the design and verification processes. As a consequence, a solid understanding of TDs is crucial.

\begin{figure}[htbp]
\centering
\includegraphics[width=0.97\columnwidth]{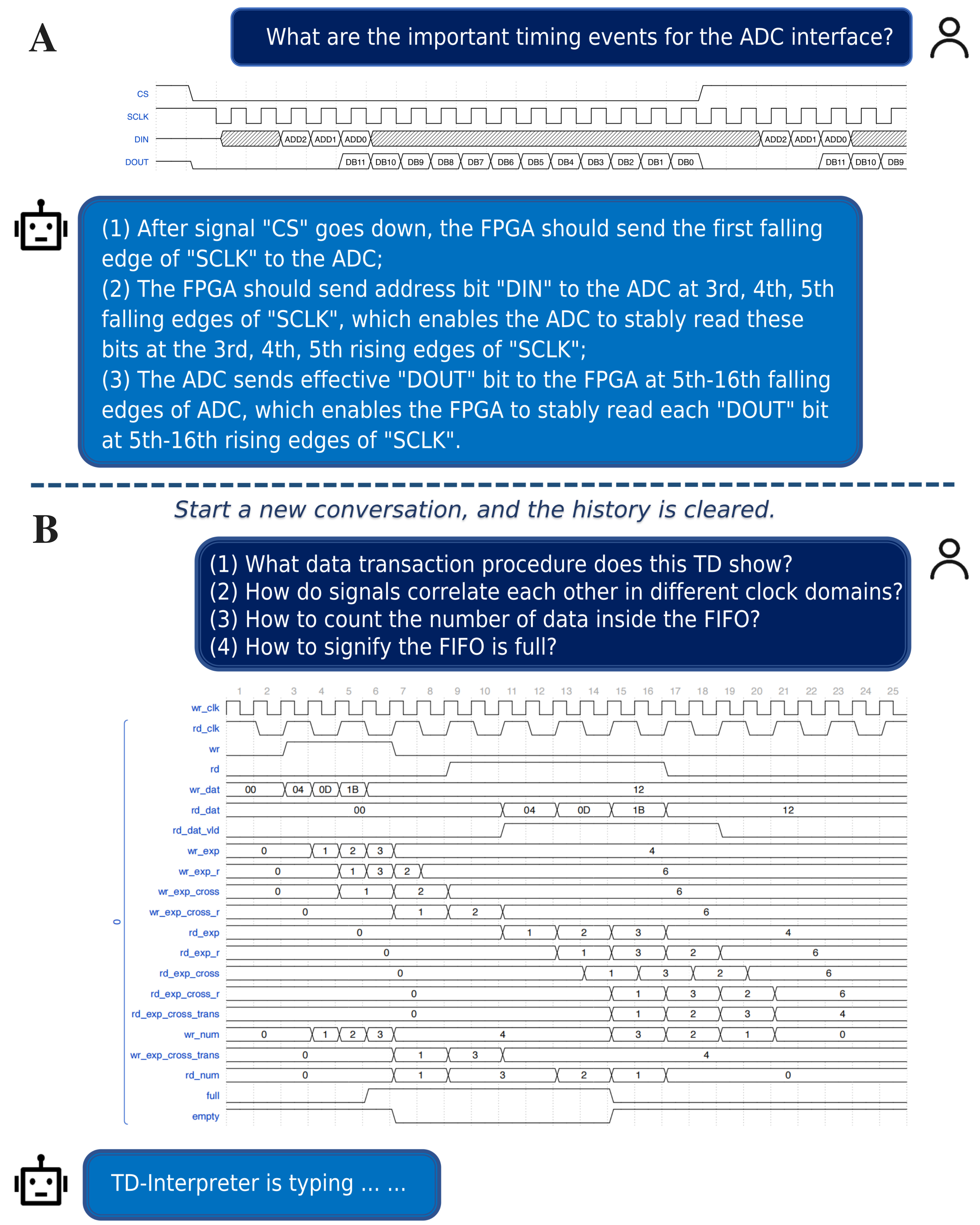}
\caption{Conversations with TD-Interpreter}
\vspace*{-3ex}
\label{fig:chatbot}
\end{figure}

Different TDs may vary in their level of abstraction. Documentations and datasheets often contain general high-level TDs describing basic timing constraints or the operational logic of specific communication protocols. TDs generated from existing code, often provide detailed insights from senior engineers. TDs drawn by engineers in daily work, are in between of these two levels. In order to help engineers better understand TDs, this paper proposes the \textit{TD-Interpreter}, an AI tool that can interpret TDs in various scenarios, as illustrated in Figure~\ref{fig:chatbot}.


Our first motivating example is understanding a set of \textit{abstract} TDs during the initial design phase. Figure~\ref{fig:chatbot}(A) shows how \textit{TD-Interpreter} can assist the engineer in designing the interface between an FPGA and an analog-to-digital converter (ADC). As shown in Figure~\ref{fig:chatbot}(A), the engineer uploads a TD picture from the datasheet\cite{adc}, and asks a question about its specification, such as, important timing events. \textit{TD-Interpreter} can then automatically analyse and interpret the TD.

Our second motivating example is understanding \textit{concrete} TDs in later design stages. Figure~\ref{fig:chatbot}(B) illustrates a use case where the engineer has to customize a third-party asynchronous first-in-first-out (FIFO) module for specialized hardware. The engineer can familiarize herself/himself with the existing module, by generating its TD pictures, and uploading them with associated questions to \textit{TD-Interpreter}.




\textit{TD-Interpreter} was developed by leveraging the advancements in multi-modal large language models (MLLMs), which have enabled diverse applications such as text-to-image generation, image captioning, and visual question-answering. However, the direct use of commercial MLLMs such as GPT-4o for TD interpretation, is still problematic due to IP-privacy concerns, fine-tuning limitations, and insufficient training data. To address these problems, we explored cost-effective solutions for data construction, training, and local deployment, specially tailored for companies with limited resources and IP concerns. In particular:

\begin{itemize}
\item\textit{We conduct a human-centered empirical study}, by interviewing a group 31 experts in embedded systems and digital design. We collected statistical information about their professional experience with TDs, the challenges and difficulties they encountered when working with TDs, their responses towards practical TDs in the wild, and their expectation regarding the functions of \textit{TD-Interpreter}.
\item\textit{We solve the lack-of-dataset problem}, by developing an automatic synthetic data generator based on the empirical study, transforming the TD-interpretation problem into a visual question-answering task in multimodal learning. We generate two types of data. The first type focuses on generating captions for given TDs, enhancing the MLLM's ability to describe detailed information within a TD. The second type targets analytical tasks, enabling the MLLM to reason about diverse timing relations.
\item\textit{We solve the IP and fine-tuning problems}, by successfully fine-tuning the open-source MLLM LLaVA~\cite{liu2023visualinstructiontuning}, with a caption-enhanced visual-reasoning approach. By combining captioning and reasoning tasks during training, LLaVA achieves very promising capabilities in reasoning about detailed information within TDs. With only 7B parameters, our model is also well-suited for local deployment.
\end{itemize}


\noindent{}In a nutshell, \textit{TD-Interpreter} combines the expertise in digital design and advances in multi-modal learning. It promotes the automatic analysis of TDs, and particularly, it digitizes the tacit knowledge in the industry, thereby accelerating the overall development process.



The organization of this paper is as follows. Section~\ref{sec:realated} provides the related work. Section~\ref{sec:questionnaire} shows details of our empirical study. Section~\ref{sec:dataset} illustrates how we generated the training data, and Section~\ref{sec:method} explores how we fine-tune the multi-modal LLM. Section~\ref{sec:evaluate} introduces and analyzes our experimental results.


\section{Related Work}
\label{sec:realated}

\noindent Since LLM's advent, multiple studies explored their use in facilitating hardware design. For example,~\cite{liu2023verilogeval, liu2023chipnemo, thakur2023benchmarking, thakur2024verigen, iccad} studied the translation of textual design requirements into Verilog code. These works either used prompt engineering in commercial LLMs, or fine-tuned open-source LLMs such as Llama~\cite{touvron2023llama} and CodeGen~\cite{nijkamp2022codegen}. Some works also extended the use of LLMs in other domains, such as accelerator design~\cite{fu2023gpt4aigchip}. However, all these works tried to convert the given problems into a natural-language-processing (NLP) code-generation task. Hence, the LLMs used by most of these works, only process textual data. However, the VQA task proposed in this paper, explores the potential of MLLM through visual-language learning, which enables the \textit{TD-Interpreter to process data in both visual and textual modalities}. Recently,~\cite{chang2024natural} also considered MLLM for hardware design, but this work did not focus on TDs, and the purpose was still code generation. 

\paragraph{Timing diagrams} 
In a recent work~\cite{he2023td}, the authors introduced \textit{TD-Magic} that can extract Strict Partial Order (SPO) relations as formal specifications from TD images. However, since only a tiny fraction of TDs contain explicit SPO relations, \textit{TD-Magic's} application is mainly restricted to creating library files for circuit components from foundries or creating timing verification scripts for tools like \textit{PrimeTime}~\cite{Bhatnagar2002}. Additionally, the multi-module, rule-based design for \textit{TD-Magic} limits its capabilities to process large-scale TD pictures. In contrast, \textit{TD-Interpreter} leverages MLLM to assist a broader range of engineers in design and verification, enabling them to interpret more complex TDs through a visual question-answer environment. However, what \textit{TD-Magic} and \textit{TD-Interpreter} have in common is that they both simulate the process of an engineer reading a TD image in digital circuits, which distinguishes them from another line of studies that directly analyze data points from possibly mixed-signal waveforms for testing or monitoring purposes~\cite{maler2004monitoring}. These works adopted a model-to-model approach to transform the data points into formal languages such as automata~\cite{fisler1999timing} and temporal logic~\cite{amla1999efficient, bartocci2022survey}. 

\paragraph{Multi-modal LLM}  MLLMs can process information from several data modalities such as text, images, or audio, and providing outputs in others~\cite{chen2024internvl}. 
As one of the most popular open-source MLLMs, LLaVA \cite{liu2023visualinstructiontuning} employs a Vision Transformer (ViT)-based vision encoder \cite{DosovitskiyB0WZ21}, a LLaMA-based language model~\cite{touvron2023llama}, and a simple linear layer as an adapter to reduce training parameters. Recently, MLLMs have been enhanced through the use of larger and more powerful text encoders or adapters~\cite{chen2024internvl}, as well as improved alignment methods~\cite{team2023gemini}. MLLMs have been demonstrated to be powerful tools for performing multimodal tasks such as image captioning, visual question answering, and visual reasoning. However, applying them to domain-specific images, such as technical diagrams, remains a challenge. In this paper, we address this issue and demonstrate that lightweight MLLMs like LLaVA can function as robust interpreters for complex tasks.
\section{Human-Centered Empirical Analysis}
\label{sec:questionnaire}
To ensure our system design reflects real-world challenges encountered by engineers, we conducted a human-centered empirical analysis on 31 professional participants regarding the use of AI tools for interpreting timing diagrams. We provide the results as follows.

\subsection{Background of Participants}
Most of the participants come from universities, research institutes, or hardware companies. Among all participants, 3 (9.7\%) have a bachelor's degree, 11 (35.5\%) have a master's degree. 17 (54.8\%) people hold a PhD's degree or during PhD study, among which there are three professors in computer engineering.

These participants cover a wide range of professional expertise in hardware design, such as digital circuit design, FPGA development, ASIC development, hardware verification, and embedded systems.

\subsection{Experience of Timing Diagrams}
We asked the participants in what contexts they typically use timing diagrams (TDs), and interestingly we found a diverse of of use cases. For example, more than half of the participants (18, 58.1\%) mentioned they used TDs to understand communication protocols (e.g., SPI, I2C, UART). Ten people (32.2\%) emphasized that they relied on TDs in debugging new hardware designs, or understanding and even reverse engineering existing designs from their colleagues or a third party. Four people (12.9\%) working in hardware verification used TDs particularly in writing/verifying Verilog/VHDL testbenches. Eight (25.8\%) highlighted the importance of using TDs in communicating with other people, such as presenting designs, group discussions, and team work. Two thirds of the participants told us they used TDs in their junior career period, such as coursework/homework assignment, or industrial internship projects. These findings are consistent with our motivation for interpreting TDs in Section~\ref{sec:intro}.

We further asked the participants \textit{Where do they usually encounter TDs}, and their responses can be divided into two groups, similar to the motivating examples in Section~\ref{sec:intro}. From their answers, half of the resources are \textit{abstract TDs} from datasheets, internal design documents, or textbooks; while the rest half are \textit{concrete TDs} waveforms from Verilog/VHDL codes, or MATLAB/Simulink simulations.

As it is important to decide which format of TDs we will support, we specially designed a question to ask the participants \textit{What tools do they use to create or analyze timing TDs}. Interestingly, we still received two groups of evenly distributed answers. The first group of people mentioned ``they mainly manually drew TDs''. The second group of resources are waveforms from simulation tools such as GTKWave and ModelSim. Despite that the participants mentioned different TD formats, in industry people tend to use standard formats for the convenience of communication, and among them \textit{wavedrom}~\cite{wavedrom-cli} is commonly used. As will be introduced in Section~\ref{sec:dataset}, \textit{wavedrom} supports the use of scripts to draw standard TDs in any level of complexity. It also provides interfaces to directly transform simulation waveforms to its standard format. Given these advantages, we chose to interpret TDs represented in the \textit{wavedrom} format.

\subsection{Challenges and Difficulties}
In this group of questions, participants were asked to use a scale from 1 (very easy) to 5 (very difficult) to rate the common challenges in understanding TDs. In the following, we show the challenges and the associated average scores of difficulty.
\begin{description}
\item[C1:] Extracting meaningful information from datasheets. (2.97/5.0)
\item[C2:] Understanding the overall transaction procedure of a communication protocol. \textbf{(3.16/5.0)}
\item[C3:] Understanding the Input/Output relations between signals. (2.84/5.0)
\item[C4:] Extracting key events from signal behavior, e.g., what a certain rising/falling edge means. (2.77/5.0)
\item[C5:] Understanding relations between multiple signals over multiple clock cycles. \textbf{(3.68/5.0)}
\item[C6:] Recognizing timing relations/constraints in multi-clock systems. \textbf{(3.77/5.0)}
\item[C7:] Analyzing clock-domain crossing behavior. \textbf{(3.84/5.0)}
\item[C8:] Measuring timing parameters (e.g., delay and latency) and identifying setup and hold time violations. \textbf{(3.26/5.0)}
\item[C9:] Detecting metastability related problems. \textbf{(4.0/5.0)}
\item[C10:] Matching timing diagrams with real-world hardware behavior. \textbf{(3.23/5.0)}
\end{description}

The challenges with average scores above 3 are considered relatively more difficult for engineers, which aligns with common experiences in digital design. In particular, understanding timing relationships across multiple signals and clock cycles, as well as identifying metastability-related issues in clock-domain crossing (CDC) scenarios, are widely recognized as complex and often intractable tasks. 

Except the difficulties listed above, several participants also added that it is challenging to extract finite state machines from waveform traces, which is a common task in analyzing communication protocols. These findings will guide our procedure in generating training data, which will be detailed in Section~\ref{sec:dataset}.

\subsection{Use of AI}
In this subsection, we investigated participants’ attitudes toward the development of \textit{TD-Interpreter} as a potential tool to assist in interpreting timing diagrams in their work. The majority of participants (30/31) indicated that they would feel comfortable using such a tool if it existed, and 24 participants expressed clear interest in its application. Over 90\% of the participants were expecting to see \textit{reasoning} procedure when \textit{TD-Interpreter} is answering questions.

Furthermore, when asked to envision potential use cases for \textit{TD-Interpreter}, participants suggested several helpful features, which are summarized below:
\begin{itemize}
    \item Automatic interpretation of timing diagrams (21, 67.7\%);
    \item Explaining step-by-step transitions in the diagram (22, 71\%);
    \item Comparing multiple timing diagrams (17, 54.8\%);
    \item Explaining ambiguous notations (16, 51.6\%);
    \item Checking for timing violations (18, 58.1\%);
    \item Generating design and verification suggestions (8, 25.8\%);
    \item Providing Verilog code recommendations based on timing diagrams (9, 29\%).
\end{itemize}

Finally, we asked participants about their expectations for the first demo version of \textit{TD-Interpreter}. The top three anticipated applications are: (1) Helping with solving problems in the HDLBits website~\cite{hdlbits}, primarily because it is a well-known platform for practicing hardware design skills; (2) Answering questions related to TDs of common peripheral protocols such as UART, I2C, SPI, mainly because these protocols are widely used by junior hardware designer, and are common in embedded systems development; (3) Answering questions related to TDs of bus protocols such as AHB or AXI, as these protocols are typically encountered by more experienced engineers and involve a rich set of timing relationships. 

Moreover, two senior participants noted that \textit{TD-Interpreter} would be more helpful if it could synthesize state machines from waveforms and identify potential issues related to clock-domain crossings (CDC).

\section{Dataset Construction}
\label{sec:dataset}

\begin{figure}[h]
    \centering
    \includegraphics[width=0.5\textwidth]{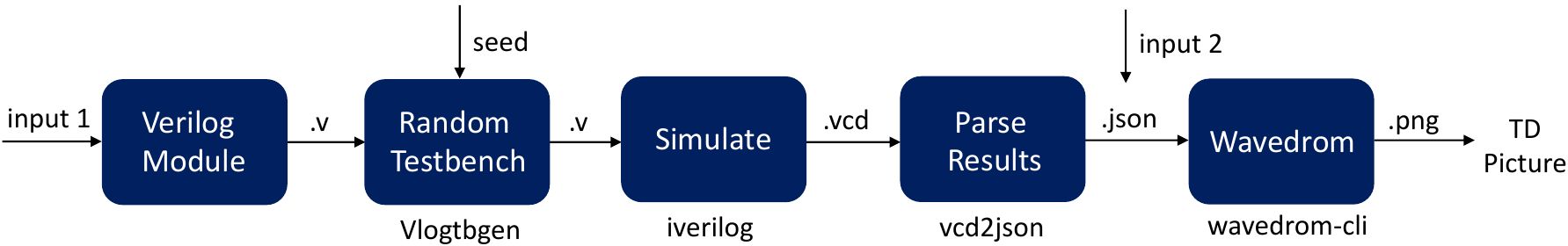}
    \caption{Workflow of Generating TD Pictures}
    \label{fig:gen_workflow}
\end{figure}

Fine-tuning an MLLM for interpreting TDs, requires a dataset of annotated TDs. However, no such a dataset currently exists. Thus we developed a strategy for generating a large dataset of TDs paired with question-answer (QA) pairs for visual instruction tuning. 

Figure~\ref{fig:gen_workflow} shows our workflow to generate TD pictures. The first path is to obtain a picture of a \textit{concrete TD}. This path starts with ``input 1'' in Figure~\ref{fig:gen_workflow}, where a Verilog module is parsed to extract basic information, such as its input and output ports. This information is then passed to the Testbench Generator \textit{Vlogtbgen} from EDAUtils~\cite{vlogtbgen}, to generate a testbench providing random input values to the module's ports. The generated testbench is then compiled and simulated using \textit{iverilog}~\cite{iverilog}, resulting in a \textrm{.vcd} file containing the simulation results. 
We use \textit{vcd2json}~\cite{vcd2json}, a tool to parse the \textrm{.vcd} file and obtain a \textrm{.json} dictionary, which includes all the necessary graphical information within the TD. Finally, the \textit{wavedrom-cli}~\cite{wavedrom-cli} tool takes this \textrm{.json} file and creates a TD picture in a \textit{wavedrom} format. This diagrammatic format is widely used in digital design. For generating a picture of an \textit{abstract TD}, we directly start from ``input 2'' by manually providing the \textrm{.json} contents from TDs in datasheets.

Our approach for data generation incorporates two phases. In the first phase, we focus on generating caption-based data. These data are centered on providing a description for a given TD picture, aiming to help MLLM to capture the details inside the TD. The second phase focuses on generating reasoning-based data. These data are based on practical use cases, aiming to enhance MLLM's ability in reasoning about diverse timing relations inside the TD.

\subsection{Caption-based Data Generation}
\subsubsection{Approach}
The strategy shown in Figure~\ref{fig:gen_workflow} allowed us to generate TDs from Verilog modules without manual input. Applying its associated process on publicly available datasets such as~\cite{verilogdataset}, supports the rapid generation of TDs. What about the QA pairs?

The simulation results of the \textrm{.vcd} file can be used to automatically create relatively simple QA pairs, which are suitable for training an MLLM to understand the basics of TDs. For instance, the \textrm{.vcd} file contains for each signal $s$, its value at clock-cycle $n$. Parsing this data allows the automatic creation of the QA pair "\textit{What is the value of signal $s$ at clock cycle $n$?}". Additional QA pairs that can be generated are, for example, "\textit{What is the sequence of values for the signal \textit{s}?}", "\textit{How many transitions does the signal \textit{s} have?}", and "\textit{How many rising edges does the signal \textit{s} have?}". These questions emphasize a basic understanding of signals' temporal characteristics, offering a solid foundation for more complex QA pairs.

More advanced QA pairs are generally difficult to programmatically generate. However, for certain Verilog modules, we successfully generated more abstract QA pairs such as "\textit{Provide a description of what you see in this TD}". This was possible because some of the modules have highly descriptive names. By leveraging these descriptive names and the available information on the module's ports, we utilized DeepSeek-Coder-V2~\cite{deepseekai2024deepseekcoderv2breakingbarrierclosedsource}, an LLM specifically designed for code-related tasks with support for Verilog code.

Using the generated-knowledge-prompting technique~\cite{0010LLWWBCH22}, the DeepSeek-Coder-V2 builds a context around a provided Verilog module. This is achieved through chain-of-thought prompting~\cite{3600270.3602070}, where the LLM is guided by a set of sub-tasks with which it can analyse the provided module in small steps. The following prompt was used to structure this chain-of-thought-prompting process: 

"\textit{You are tasked with providing brief descriptions of various digital components based on limited data. For each one, provide a description of the component. To do this perform the following steps: 1. Assume it is a Verilog module. 2. Split the module's name by the '\_' character. 3. List and analyse each individual part of the module name. 4. List and count the input and output ports. 5. Analyse the names of the input and output ports. 6.~Determine the possible functionality of each of the ports. 7.~Determine the functionality of the module based on the individual parts of the name, and the I/O ports.}". 

Following this prompt, the LLM was able to provide clear and detailed responses, to prompts asking it to provide a description, a summary, and a list of use cases, for the provided Verilog module.

\begin{figure}[h]
    \centering
    \includegraphics[width=0.48\textwidth]{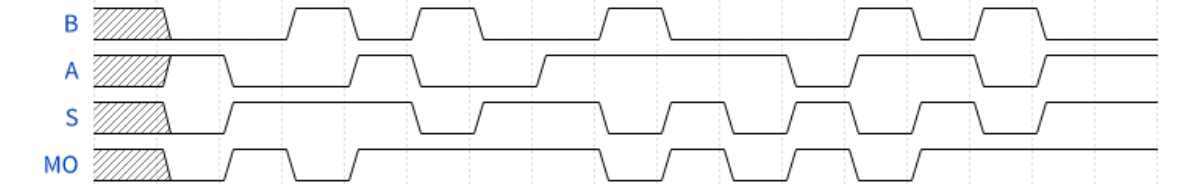}
    \caption{TD of a 2-to-1 negative multiplexer}
    \label{fig:2-1-negative-mux}
    \vspace*{-2ex}
\end{figure}

An example of our training strategy, is the outcome obtained for the TD shown in Figure \ref{fig:2-1-negative-mux}. This represents a Verilog module named "\textit{maxv\_nmux21}". Through the above-described prompting process, the DeepSeek-Coder-V2 LLM successfully identified the module as a 2-to-1 negative multiplexer. It thus provided a detailed description, where it explains its purpose and its input and output ports, a fitting caption for the TD, a brief summary of the TD, and a list of use cases of the module. To highlight the need for such training data, we provided this TD to GPT-4o with a prompt asking for explanation. It only responded with a basic enumeration of the signals, and a generic explanation of how they switch between high and low. It also misidentified the A and B input signals as clock signals, and it misinterpreted the diagram as potentially representing some sort of clock signal relationship or a state machine.

\begin{figure}[h]
    \centering
     \vspace*{-2ex}
    \includegraphics[width=0.48\textwidth]{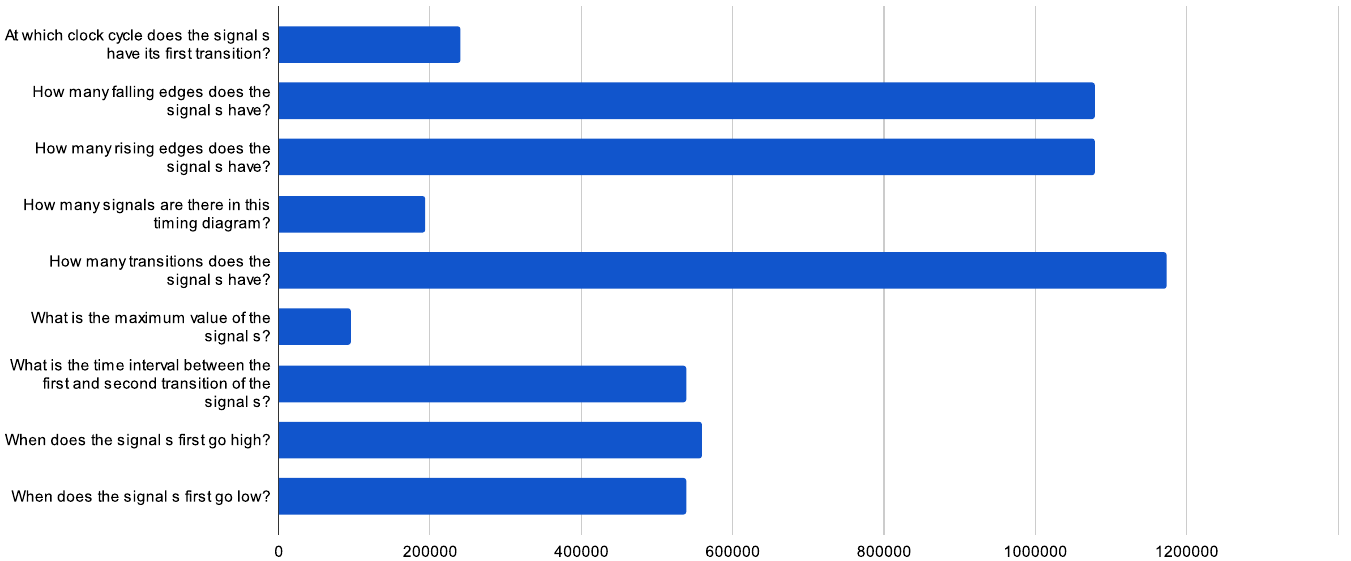}
    \caption{Question Types Distribution}
    \label{fig:qa-pairs-chart}
    \vspace*{-1ex}
\end{figure}

\subsubsection{Statistics}
Through the outlined caption-based data-generation method, a total of 221,983 TDs were generated. For these TDs, we were able to generate 9,915,694 QA pairs, of which 4,306,008 are of the form "\textit{What is the value of signal s at clock cycle n?}", which requires an MLLM to detect concrete signal values at specific time instances. The distribution of the other questions are shown in Figure \ref{fig:qa-pairs-chart}. In addition, for 28,271 of the TDs, the corresponding Verilog module's name was sufficiently descriptive enough to generate additional captions using DeepSeek-Coder-V2. For these TDs we were also able to generate the following questions: "\textit{Provide a description/caption/summary/use case of the TD}".


\subsection{Reasoning-based Data Generation}
\label{subsec:reasoning_data_generate}
Unlike caption-based data, which can be partially automated with EDA tools, generating reasoning-based data relies heavily on manual inspection of our empirical analysis in Section~\ref{sec:questionnaire}, including the selection of use cases and the reflection of typical challenges engineers face when working with TDs. 

We identified a set of commonly used components in the system designs. They include memory, communication protocols, and timers. We also considered design problems from the HDLBits website~\cite{hdlbits}, which are widely recognized benchmarks to evaluate the coding generation capability of LLMs.
For each component, we manually went through every design step, and particularly, we paid a special attention on some critical timing points where designers can easily make subtle mistakes. We also intentionally crafted very specific and even unconventional questions, which require solutions depending highly on experience and the understanding of the concrete context.

To demonstrate how we generated QA pairs from a \textit{concrete TD} picture, we first use the design of a serial receiver protocol~\cite{serialprotocol}. This is an illustrative example, due to its complexity. The same methods can be extended to other use cases. We then briefly discussed the approach in generating data for \textit{abstract TDs}, which is simpler. We also provide a statistical analysis for the generated data.

\begin{figure}[h]
    \centering
    \includegraphics[width=0.48\textwidth]{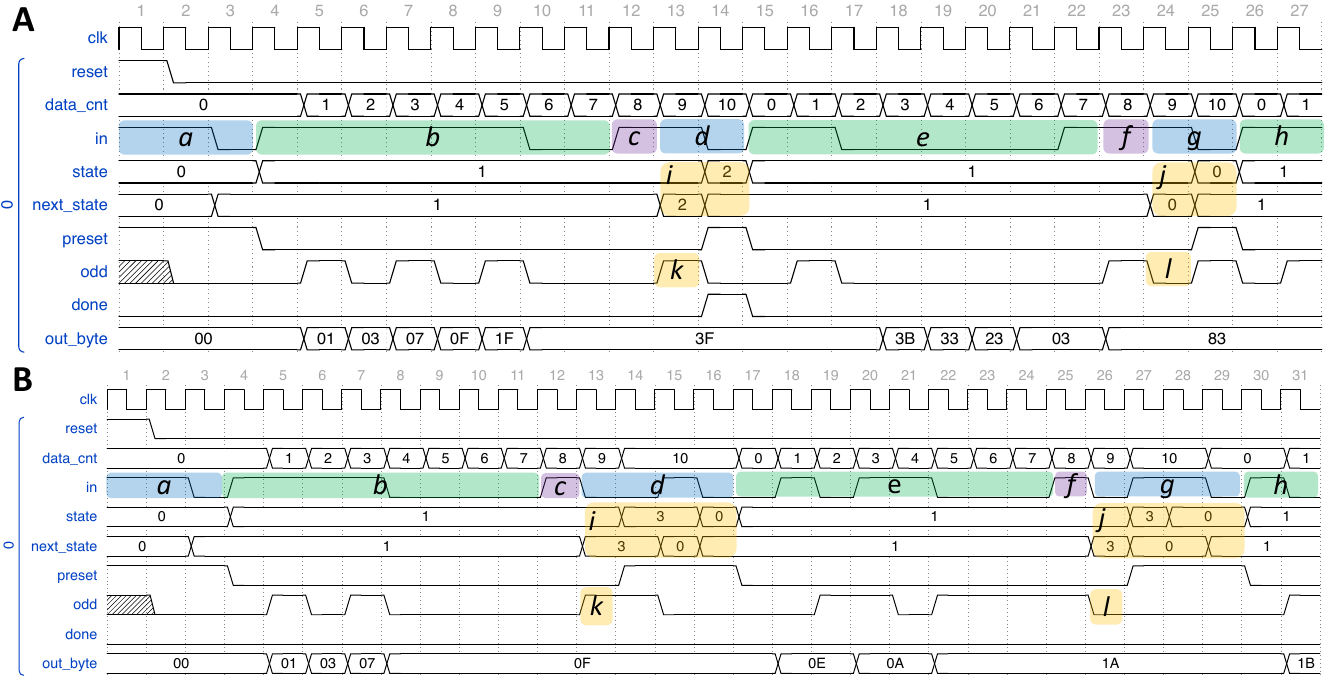}
    \caption{Desired TDs for the Serial Receiver Protocol}
    \label{fig:serial-td}
\end{figure}

\begin{figure}[h]
    \centering
    \vspace*{-4ex}
    \includegraphics[width=0.34\textwidth]{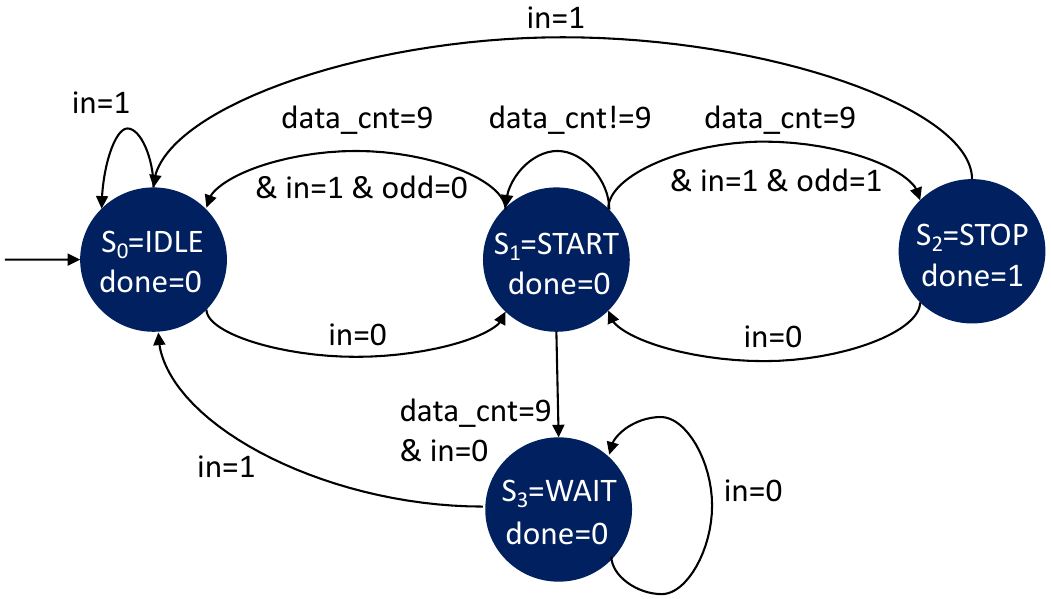}
    \caption{Moore FSM for the Serial Receiver}
    \label{fig:serial-fsm}
    \vspace*{-2ex}
\end{figure}

\subsubsection{Generate QA pairs}
The \textit{Serial Receiver} example describes a communication protocol that receives a stream of bytes with the extra 9th bit and 10th bit for parity and stop checking respectively. The requirements for correctly receiving a byte are: \textbf{(R1)} The first 9 bits should have odd number of 1; \textbf{(R2)} The 10th bit is 1. Besides, the engineer will often see the desired TDs as shown in Figure~\ref{fig:serial-td}. 

Accompanied with textual requirements, a typical design workflow requires the engineer to analyze and reason about the desired TDs. Our methods of constructing QA pairs also follow this procedure.

\paragraph{FSM design} Figure~\ref{fig:serial-fsm} shows a Finite State Machine (FSM) that represents the transaction logic of the aforementioned protocol. Figure~\ref{fig:serial-td} (A) shows the success scenario where the states transit between $S_0, S_1, S_2$, while Figure~\ref{fig:serial-td} (B) shows the failure situation where the states transit between $S_0, S_1, S_3$. As given by the textual requirements and the desired-TDs, building a partial/complete FSM is a classic problem in digital design, which we consider in particular.

\paragraph{Datapath design} As shown in Figure~\ref{fig:serial-fsm}, $data\_cnt=9$ is a critical condition for state transitions, hence the design of counter signals like $data\_cnt$ is very important. Given the desired-TD, we may ask: \textit{In which scenarios will ``data\_cnt'' become 0, keep unchanged, and add 1?} Additionally, as shown in  Figure~\ref{fig:serial-td}, signal $out\_byte$ is closely related to the sequence in signal $in$. We can also ask: \textit{How does ``out\_byte'' store the bits from ``in'' sequentially}?

\paragraph{Critical timing analysis} Since engineers are often confused about specific behaviors of the waveform, this type of questions are specially designed for this scenario. They can also be used to enhance MLLM's sensitivity in combinatorial logic and sequential logic. For example, given the proposed FSM in Figure~\ref{fig:serial-fsm}, we can ask questions about the event when the protocol starts to receive the first data stream: \textit{What happens at the 3rd clk rising edge and the following one clk cycle?} The answer would provide an analysis from the FSM:  \textit{At the 3rd rising edge, state=$S_0$. However, ``in=0'' in the subsequent clk cycle, which triggers ``next\_state'' to become $S_1$ through combinatorial logic.  Then in the 4th clk rising edge, ``state'' will get sampled as $S_1$ through sequential logic.} Similarly, we can ask what happens when the first data stream of date finishes transmission: \textit{Please give an explanation of state transition when data\_cnt=9.}

Apart from the type of questions, we also considered their formats. For example, we changed some statement questions into True/False form, or a multiple choice form to enrich the diversity of questions.

\subsubsection{Generate Concrete TDs}
As the state transition of the FSM relies heavily on the input signal $in$, a purely random approach to generate TDs will result in unpredictable outputs, for which it is difficult to assign QA pairs. To address this challenge, we developed a \textit{top-down-conditioned-random-generation} approach, balancing between the variety of TDs, and the convenience of designing QA pairs.

\paragraph{Scenario division} This procedure divides the transaction procedure into different scenarios. As shown in Figure~\ref{fig:serial-td}, we group the traces of the FSM into two groups: success (A) and failure (B).

\paragraph{Clamp insertion} Before randomly assigning inputs, we add fixed clamps in the input sequence, as shown by Bands $a, c, d, f, g$ in both Figure~\ref{fig:serial-td} (A) and (B). Band $a$ represents a ``110'' sequence to ensure that $state$ will become $S_1$ at the 4th clk rising edge to start bit transmission. We assign 1 in Bands $c, f$ for parity checking.

\paragraph{Random generation} Bands $b, e, h$ in Figure~\ref{fig:serial-td} (A) and (B) are where we assign conditioned random inputs. For example, for Band $b$ in Figure~\ref{fig:serial-td} (A) and (B), the sequence must have even number of 1s, which makes the number of 1s in Bands $b+c$ become odd. Conversely, we require the sequence in Band  $e$ to have odd number of 1s, which makes the number of 1s in Bands $e+f$ become even.

\paragraph{Motif reproduction} With the clamps and the conditioned random inputs, we can reproduce fixed motifs in the TDs. For example, as the number of 1s in Bands $b+c$ is odd, we can observe a high pulse of $odd$ in Band $k$ from both Figure~\ref{fig:serial-td} (A) and (B). In contrast, the value of $odd$ in Band $l$ is low due to the even number of 1s detected. Moreover, in Figure~\ref{fig:serial-td} (A) and (B), the chain reactions of of Bands $d, k$ and Bands $g, l$ will respectively cause fixed results in Band $i$ and Band $j$. The stable motif shown in Bands $i, j, k, l$ will thus facilitate the assignment of QA pairs. 
 
\subsubsection{Generate Data for Abstract TDs}
For \textit{abstract TDs}, we focused on the commonly used AMBA AXI~\cite{axi}, AHB~\cite{ahb}, and APB~\cite{apb} protocols, and also the interface design of ADCs~\cite{adc}, which are widely used in embedded systems. The manuals of these protocols contain several TDs illustrating key scenarios, such as read or write transfers in APB and burst transfers in AHB. From these resources, we selected several TDs and manually crafted QA pairs for them. These QA pairs were specifically designed to address common engineering challenges.  Given the small scale of these \textit{abstract TDs}, we adopted a lightweight method for randomizing their presentation, including altering signal order, varying clock cycles, and changing letter case.


\subsubsection{Statistics}

\begin{table}[h]
\caption{\# Types of Questions}
\vspace{4ex}
\centering
\small
\resizebox{0.48\textwidth}{!}{
\begin{tabular}{l|c|l|c}
\hline
\textbf{Task}         & \textbf{\# Type} & \textbf{Task}         & \textbf{\# Type} \\ \hline
serial\_datapath\_stop~\cite{serialdatapath}       & 24               & motor\_timer\_failure~\cite{motor_timer}        & 13                    \\ \hline
serial\_datapath\_wait~\cite{serialdatapath}    & 24                  & complete\_fsm~\cite{complete_fsm}                & 13                    \\ \hline
serial\_parity\_stop~\cite{serialprotocol}          & 22                    & fancy\_timer~\cite{fancy_timer}                   & 16                    \\ \hline
serial\_parity\_wait~\cite{serialprotocol}         & 22                    & sync\_fifo                   & 15                    \\ \hline
HDLC\_correct~\cite{hdlc}                 & 13                    & async\_fifo                        & 14                    \\ \hline
HDLC\_discard~\cite{hdlc}                  & 12                    & spi\_no\_clk                 & 25                    \\ \hline
HDLC\_error~\cite{hdlc}                    & 13                    & spi\_with\_clk               & 33                    \\ \hline
w\_counter~\cite{wcounter}                    & 10                    & ADC~\cite{adc}                          & 12                    \\ \hline
motor\_timer\_success~\cite{motor_timer}        & 13                    & AXI/AHB/APB~\cite{axi,ahb,apb}                  & 16                    \\ \hline
\end{tabular}
}
\label{tab:questions}
\end{table}

\begin{figure}[h]
    \centering
    \includegraphics[width=0.32\textwidth]{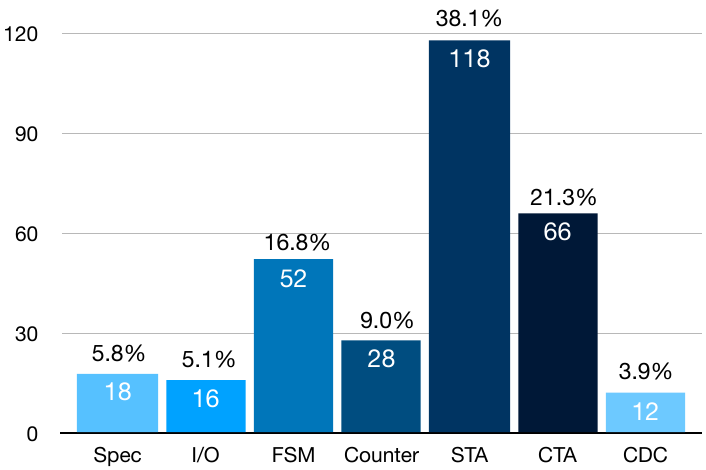}
    \caption{Question Distribution}
    \label{fig:question_distribute}
\end{figure}

Table~\ref{tab:questions} showcases all the design tasks we considered. Overall, we designed 310 different types of questions. Figure~\ref{fig:question_distribute} further shows their distribution. Some tasks have a specific type of questions. For the ``async\_fifo'' case, the mostly asked questions are related to clock-domain crossing (CDC), as shown from the second example in Figure~\ref{fig:chatbot}. We also find that most questions come from the groups of simple-timing analysis (STA) and complex-timing analysis (CTA). STA questions require an explanation of events for a single signal, or for a single clock cycle, while CTA questions require an answer about the relations across different signals. For design cases in practice, such as ``spi\_no\_clk'' (the slave has no clock) and ``spi\_with\_clk'' (the slave has an independent clock), the complexity escalates significantly, so most of the questions fall in the CTA group. In Section~\ref{subsec:exp_results} we provided concrete examples for several types of questions.

\section{The TD-Interpreter}\label{sec:method}

\subsection{Architecture}\label{subsec:model_architecture}

\begin{figure}[htbp]
\centering
\includegraphics[width=0.48\textwidth]{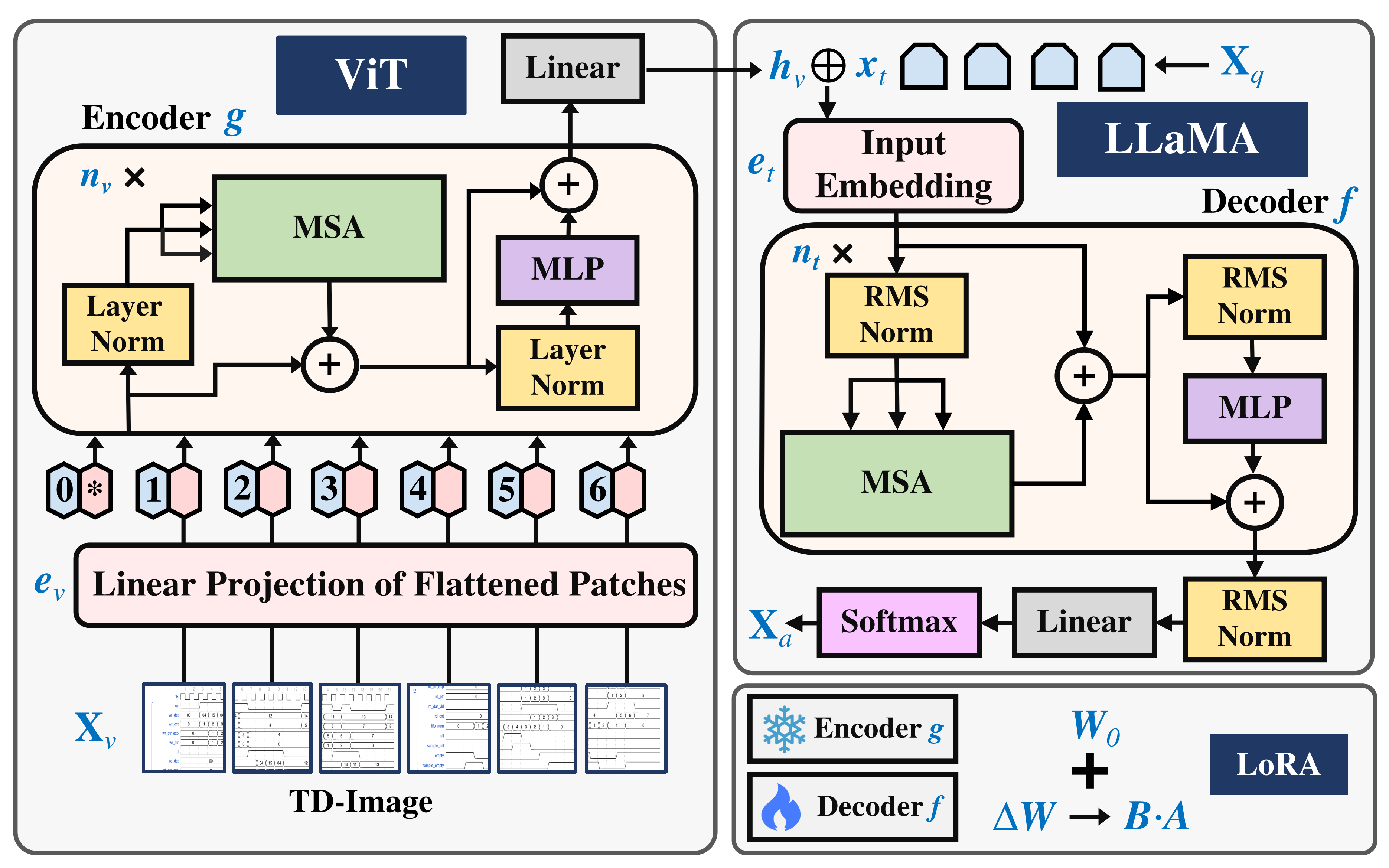}
\caption{Model Architecture of TD-Interpreter}
\label{fig:arch}
\end{figure}

As shown in Figure~\ref{fig:arch}, \textit{TD-Interpreter} is a fine-tuned LLaVA model~\cite{liu2023visualinstructiontuning} developed using our dataset. It incorporates a 
vision encoder $g(\cdot)$ based on the vision transformer (ViT)~\cite{dosovitskiy2020image}, which enables the model to comprehend image information, alongside a language model $f(\cdot)$ that generates textual responses based on both the visual input and the textual request. \textit{TD-Interpreter} takes an image of a TD $\mathbf{X}_v\in \mathbb{R}^{H\times W\times C}$ and a textual question $\mathbf{X}_q$ as the inputs, where $H \times W$ represents the resolution of $\mathbf{X}_v$, and $C=3$ represents the number of RGB channels in the TD picture. The vision encoder $g(\cdot)$ first extracts the visual features from the input image as $\mathbf{h}_v = g(\mathbf{X}_v)$. 
Subsequently, the language model predicts the next token of the target textual response $\mathbf{X}_a$ from the input embedding $\mathbf{e}_t$ based on the visual embedding $\mathbf{h}_v$ and the textual input sequence of tokens $\mathbf{x}_t$ converted from $\mathbf{X}_q$, represented as $\mathbf{e}_t = f(\mathbf{h}_v, \mathbf{x}_t)$. 

\subsubsection{Vision Encoder}\label{subsec:vision_encoder}
The ViT-based~\cite{dosovitskiy2020image} vision encoder $g(\cdot)$ consists of a sequential stack of $n_v$ identical transformer blocks $g_i(\cdot), 1 \le i\le n_v$, along with an input embedding layer $e_v(\cdot)$, and an output linear layer. Each block includes a multi-head self-attention (MSA) layer, a multi-layer-perceptron (MLP) layer, and two Layernorm (LN)~\cite{ba2016layer} layers. The detailed encoding of the input TD image $\mathbf{X}_v$ can be represented as follows.
First, the image $\mathbf{X}_v$ is patchified into a sequence of flattened 2D patches represented as 
$\mathbf{x}_v \in\mathbb{R}^{N\times (P^2 \cdot C)}$, where $N=(H\times W)/P^2$ is the number of patches, and $P \times P$ indicates the resolution of each patch. Then, the sequence of patches is encoded by the input layer $e_v$ as follows: $\mathbf{z}_p^0 = e_v(\mathbf{x}_v)$.

Subsequently, the sequence $\mathbf{z}_p^0$ is processed by the transformer blocks $g_i(\cdot), 1 \le i\le n_v$, where the encoding procedure of the $i$-th block $\mathbf{z}_p^{i} = g_i(\mathbf{z}_p^{i-1})$ can be denoted as follows: $\hat{\mathbf{z}}_p^{i-1} = \text{MSA}(\text{LN}(\mathbf{z}_p^{i-1})) + \mathbf{z}_p^{i-1}$, and $\mathbf{z}_p^{i} = \text{MLP}(\text{LN}(\hat{\mathbf{z}}_p^{i-1})) + \hat{\mathbf{z}}_p^{i-1}$.

Finally, we get the visual feature $\mathbf{h}_v = \text{Linear}(\mathbf{z}_p^{n_v})$ by transforming $\mathbf{z}_p^{n_v}$ using the last linear layer of the vision encoder.

\subsubsection{Language Model}\label{subsec:language_model}
The language model $f(\cdot)$ is a pre-trained large-language model (LLM) based on the LLaMA architecture~\cite{touvron2023llama}, parameterized by a Vicuna~\cite{chiang2023vicuna} model as described in~\cite{liu2023visualinstructiontuning}. Specifically, $f(\cdot)$ consists of an input embedding layer $e_t(\cdot)$, $n_t$ identical transformer blocks $f_i(\cdot), 1\le i \le n_t$, and an output block comprising an RMSNorm (RN)~\cite{zhang2019root} layer and a linear layer. Given an textual input sequence of tokens $\mathbf{x}_t$ converted from the request $\mathbf{X}_q$, it is first encoded by the input layer $e_t(\cdot)$ as follows: $\mathbf{z}_t^0 = e_t(\text{Concatenate}(\mathbf{x}_t, \mathbf{h}_v))$.

Then, $\mathbf{z}_t^0$ is encoded by $n_t$ transformer blocks, where the encoding procedure for the $i$-th block $\mathbf{z}_t^{i} = f_i(\mathbf{z}_t^{i-1})$ can be represented by $\hat{\mathbf{z}}_t^{i-1} = \text{MSA}(\text{RN}(\mathbf{z}_t^{i-1}))+\mathbf{z}_t^{i-1}$ and $\mathbf{z}_t^i = \text{MLP}(\text{RN}(\hat{\mathbf{z}}_t^{i-1})) + \hat{\mathbf{z}}_t^{i-1}$.

Finally, the predicted logits $\mathbf{h}_t$ can be expressed by $\mathbf{h}_t = \text{Linear}(\text{RN}(\mathbf{z}_t^{n_t}))$. By employing the \textit{softmax} on $\mathbf{h}_t$ and extracting the most likely token in each decoding step, we get the predicted next token of target textual response $\mathbf{X}_a$.

\subsection{Training}\label{subsec:training}
As detailed in Section~\ref{sec:dataset}, for each TD image $\mathbf{X}_v$, we construct a single-turn conversation data $(\mathbf{X}_q, \mathbf{X}_a)$. Then, for a target answer sequence $\mathbf{X}_a$ of length $l$, we compute the probability of it by: $p(\mathbf{X}_a|\mathbf{X}_v, \mathbf{X}_q) = \prod_{i=1}^l p_\theta(x_i|\mathbf{X}_v, \mathbf{X}_{q}, \mathbf{X}_{a,<i})$.

Here, we fix the parameters of the vision encoder $g(\cdot)$ while leaving the parameters of the language model $f(\cdot)$ trainable, denoted by $\theta$. Additionally, $\mathbf{X}_{q,<i}$ and $\mathbf{X}_{a,<i}$ represent the tokens of the question and the generated answer before the the current $i$-th token.

To enhance the speed and efficiency of training, we employ the low-rank-adaptation (LoRA) technique~\cite{hu2021lora}, on all layers of our TD-interpreter, and conduct fine-tuning. Intuitively, LoRA introduces a trainable adapter that contains significantly fewer parameters than the original weights. By training only the lightweight adapter instead of the full weight matrix, LoRA substantially alleviates the training burden.
Specifically, given a pre-trained weight matrix $W_0 \in \mathbb{R}^{d\times k}$, where $d\times k$ is the shape of the matrix, LoRA modifies the weight matrix from $W_0$ to $W_0 + \Delta W = W_0+B \cdot A$, where $A\in \mathbb{R}^{r\times k}$ and $B\in \mathbb{R}^{d\times r}$ are two trainable low-rank matrices with $r \ll \min(d, k)$. This technique effectively reduces the number of trainable parameters for each layer from $d \times k$ to $r \times (d + k)$.

\section{Experimental Evaluation}
\label{sec:evaluate}

\subsection{Training details.} 
We first prepared an evaluation dataset as a benchmark for testing. This dataset covered almost all use cases in Table~\ref{tab:questions}. It also contained all types of questions in Figure~\ref{fig:question_distribute} that represented various cause-effect relations between signals. For each question-answer pair, due to random generation, all TD pictures looked different and were never seen during training. The most complex picture included up to 51 signals and 80 clock cycles. As illustrated in Section~\ref{subsec:reasoning_data_generate}, these data were first created by humans then automated for random generation, so evaluation on these data is able to manifest the model’s performance across different real use cases and question types.

For the training data set, although the use of our data generator described in Section~\ref{sec:dataset} is capable of generating millions of synthetic data, we found that it was not necessary to train the total amount of data, and this was not economical for institutes or companies with limited computation resources. We empirically found it would yield desirable evaluation results if the overall amount of training data reached around 10k. Consequently, for the caption-based data, we randomly sampled 4942 TD pictures and associated each picture with one QA pair; For the reasoning-based data, we randomly sampled 5292 TD pictures, with each picture a single QA pair. We then were able to train our model on 8 NVIDIA A800 GPUs with a batch size of 32. The training process takes 17 hours with 100 epochs. We utilized a learning rate of $10^{-4}$ and set the LoRA rank to 8. The LoRA adapter consists of 20M parameters.

\subsection{Experimental results.} 
\label{subsec:exp_results}
As the outputs of the Visual QA task are texts, and the ground truth is also given in form of texts, we can use commonly used metrics in text comparison, such as Bleu-x and Rouge-x to evaluate the accuracy of model outputs. ``-x'' represents different approaches in calculating the specific metrics. The range of Bleu-x and Rouge-x are between 0 and 100, while 100 means the model outputs completely overlap the labelled answers, and 0 means the worst case. In this paper we only focus on the absolute values of Bleu-x and Rouge-x.

We first evaluated the untuned LLaVA and got the following results: Bleu-4=13.9, Rouge-1/2/l=23.2/6.6/16.3. In comparison, after finetuning, LLaVA can promisingly achieve: Bleu-4=95.9, Rouge-1/2/l=96.7/95.9/96.5. The significance of these comparison results is two-fold: (1) Before finetuning, LLaVA has very limited capability in understanding TDs, and after finetuning, there was a huge improvement in its performance; (2) We observe that after finetuning, LLaVA can achieve over 95 scores across all metrics of Bleu-x and Rouge-x. This means that for all the questions in the evaluation dataset, LLaVA can answer them correctly, except only some small errors. 

In the following, we provide four illustrative examples showing the capabilities of \textit{TD-Interpreter}. After each index of question, we added its question type that mapped to Figure~\ref{fig:question_distribute}. We selected relatively more complex questions, such as questions in groups of FSM, CDC and CTA, to showcase the reasoning capability of \textit{TD-Interpreter} in complicated timing relations. Some questions require unconventional answers as they came from specific design issues. We also compared our results with the ones of GPT-4o. Texts in blue show where GPT-4o made mistakes. Moreover, we provided human feedback to reflect the typical problems that designers may face.

\subsubsection{AHB} 
\begin{figure}[htbp]
\centering
\includegraphics[width=1\columnwidth]{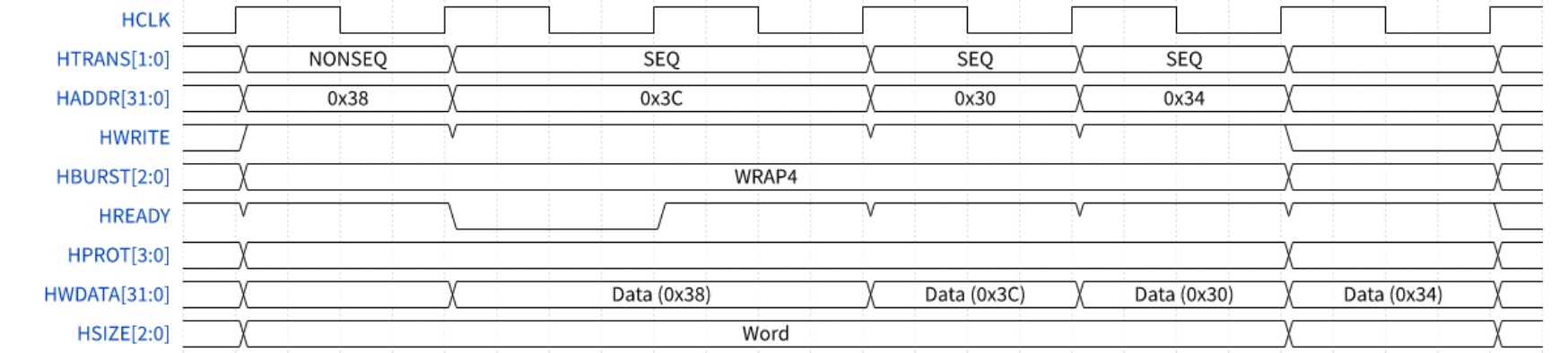}
\caption{AHB Burst Write Transfer}
\label{fig:ahb-burst-transfer}
\end{figure}

The TD in Figure \ref{fig:ahb-burst-transfer} is a burst-write-transfer example using the advanced-high-performance-bus (AHB) specification.

\begin{itemize}
    \item \textbf{Q1 (CTA):} Why takes writing to address 0x3C longer than other transfers?
    \item \textbf{A1 (TD-Interpreter):} <<{\color{blue}The write to address 0x3C does not take longer. Instead it's the write to address 0x38 which takes longer.} One of the subordinates adds a wait state in this transfer, as indicated by the HREADY signal being LOW. This extends the address phase of the transfer after it. In this case this is the write transfer to address 0x3C. {\color{blue}In summary, the data phase of the transfer to address 0x38 is including a wait state, which extends the address phase of the transfer to address 0x3C.}>>
    \item \textbf{A1 (GPT-4o):} GPT-4o notices that signal HREADY is low during the write transfer. However, it fails to see that this occurs due to the wait states inserted during the previous transfer. It then falsely identifies the write to address 0x3C as the first transfer in the burst.

    \item \textbf{Human feedbcak:} In our empirical study, five participants who claimed to be familiar with AHB protocol tried to analyze this TD. However, only 1 person correctly analyzed the timing behavior as asked in \textbf{Q1}, while most participants simply said ``don't know''. The average confidence level is also low, which is only 2.2/5 (1 means not confident, 5 means very confident).
\end{itemize}

\begin{figure}[htbp]
\centering
\includegraphics[width=1\columnwidth]{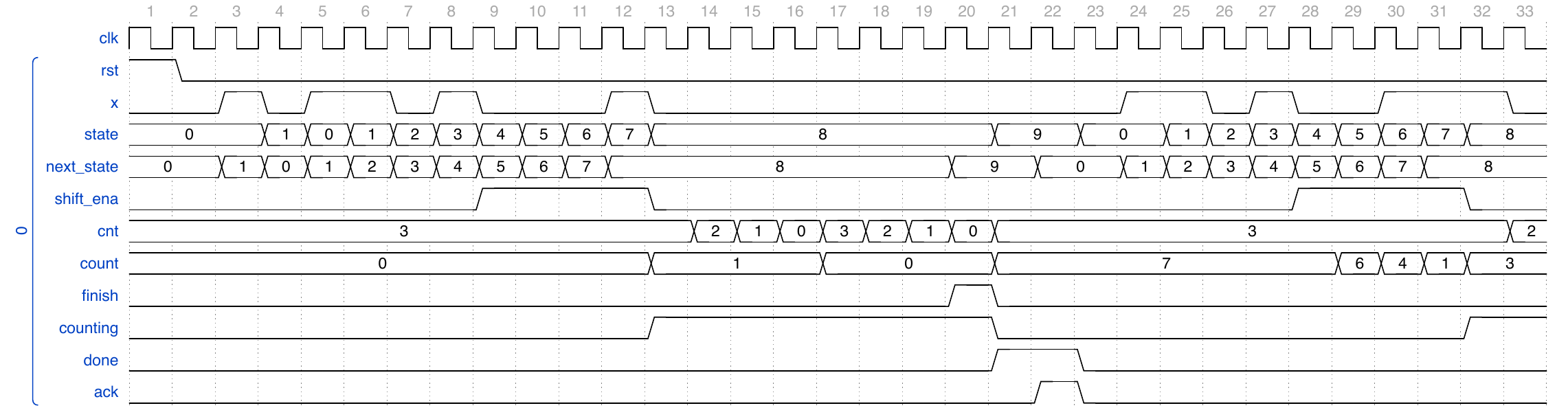}
\caption{Fancy Timer}
\label{fancyTimer}
\end{figure}

\subsubsection{Fancy Timer} The TD in Figure~\ref{fancyTimer} shows a module of a fancy timer, satisfying requirements: \textit{(R1)} Get started when a particular pattern (1101) is detected, \textit{(R2)} Shift in 4 more bits to determine the duration to delay, \textit{(R3)} Wait for the counters to finish counting, \textit{(R4)}~Notify the user and wait for the user to acknowledge the timer.
\begin{itemize}
    \item \textbf{Q1 (FSM):} Can you synthesize a Moore FSM?
    \item\textbf{A1 (TD-Interpreter):} <<{\color{blue}\textit{(R1)} $s_0$-\textit{x}->$s_1$, $s_0$-$\sim$\textit{x}->$s_0$, $s_1$-\textit{x}->$s_2$, $s_1$-$\sim$\textit{x}->$s_0$, $s_2$-\textit{x}->$s_2$, $s_2$-$\sim$\textit{x}->$s_3$, $s_3$-\textit{x}->$s_4$, $s_3$-$\sim$\textit{x}->$s_0$; \textit{(R2)} $s_4$-$\sim$\textit{rst}->$s_5$, $s_4$-\textit{rst}->$s_0$, $s_5$-$\sim$\textit{rst}->$s_6$, $s_5$-\textit{rst}->$s_0$, $s_6$-$\sim$\textit{rst}->$s_7$, $s_6$-\textit{rst}->$s_0$, $s_7$-$\sim$\textit{rst}->$s_8$, $s_7$-\textit{rst}->$s_0$; \textit{(R3)} $s_8$-\textit{finish}->$s_9$, $s_8$-$\sim$\textit{finish}->$s_8$; \textit{(R4)} $s_9$-\textit{ack}->$s_0$, $s_9$-$\sim$\textit{ack}->$s_9$; \textit{(Out)} $s_4 \sim s_7$->\textit{shift\_ena}=1, $s_8$->\textit{counting}=1, $s_9$->\textit{done}=1.}>>
    
    \item \textbf{A1 (GPT-4o):} A very high-level design of the FSM.

    \vspace*{1mm}\item \textbf{Q2 (Counter):} In which clk cycle do both ``cnt'' and ``count'' equal to 0, and why in the next clk cycle the timer immediately stops working?
    \item \textbf{A2 (TD-Interpreter)}: <<{\color{blue} Look at the 20th clk cycle. During this cycle, both ``cnt'' and ``count'' equal to 0.} This could be used as a triggering signal to set ``finish'' as 1.
    Since current state is $s_8$, and according to the FSM, if $finish=1$ in this state, ``next\_state'' will be $s_9$.
    Then at the 21st clk rising edge, ``state'' gets sampled as $s_9$. This will disable signal ``count'', and enable signal ``done''>>.

    \item \textbf{A2 (GPT-4o):} Fails to locate the correct clk cycle, where both signals ``cnt'' and ``count'' have a value equal to 0.

    \item \textbf{Human feedback:} We did not ask participants to analyze this TD as it takes time even for human experts to synthesize a FSM from this TD. However, as mentioned in Section~\ref{sec:questionnaire}, it is a practical need to generate FSMs out of waveforms, so it would save considerable time for engineers when \textit{TD-Interpreter} provides this function. 
\end{itemize}


\subsubsection{Asynchronous FIFO} The TD in Figure~\ref{fig:chatbot}(B) shows an asynchronous FIFO module with $depth=4$.
\begin{itemize}
    \item \textbf{Q1 (CDC):} Is the following relation correct? 
    
    grayencode(\textit{wr\_exp}) -> \textit{rd\_clk} -> \textit{wr\_exp\_r} -> \textit{wr\_clk} -> \textit{wr\_exp\_cross} -> \textit{wr\_clk} -> \textit{wr\_exp\_cross\_r} -> \textit{wr\_exp\_cross\_trans} = graydecode(\textit{wr\_exp\_cross\_r}).
    \item \textbf{A1 (TD-Interpreter):} <<{\color{blue} No, the correct relation should exchange \textit{rd\_clk} and \textit{wr\_clk}.}>>

    \item \textbf{A1 (GPT-4o):} Sometimes GPT-4o thought the provided relation is correct, and sometimes GPT-4o thought it is wrong, but it tended to provide an incorrect solution.

    \vspace*{1mm}\item \textbf{Q2 (CDC):} Observe the first reading operation, and answer: \textit{(1)} Does ``wr\_num'' represent the real number of data inside the FIFO? \textit{(2)} Does ``rd\_num'' represent the real number of data inside the FIFO? 
    \item \textbf{A2 (TD-Interpreter):} <<\textit{(1)} No. ``wr\_num'' works in the ``wr\_clk'' domain, so in scenarios where consecutive reading operation happens, ``wr\_num'' cannot reflect the real number of data inside FIFO.
    \textit{(2)} {\color{blue} Yes. From the TD, $rd\_num=wr\_exp\_cross\_trans-rd\_exp$. }``rd\_exp'' has no information lag in the ``rd\_clk'' domain, but ``wr\_exp\_cross\_trans'' has. However, before the reading operation, ``wr\_exp\_cross\_trans'' has already reached the correct value 4, and remains unchanged during reading. {\color{blue} Hence, ``wr\_exp\_cross\_trans'' is correct in this case.}>>  

    \item \textbf{A2 (GPT-4o):} In most cases ``rd\_num'' cannot reflect the number of data inside the FIFO, but the particular case shown in \textit{Asynchronous FIFO} is an exception. GPT-4o simply provided a negative answer with textbook-style explanations, and the way it proposed to compute ``rd\_num'' was wrong.

    \item \textbf{Human feedback:} Six participants in our empirical study tried to analyze this TD. First, they all agreed that they may ask the questions in Figure~\ref{fig:chatbot}(B) to gain a basic understanding of this asynchronous FIFO module. However, when it came to questions about identifying clock domain crossing (CDC) issues, these participants made mistakes, despite that FIFO was known to be basic for digital designers. For example, in \textbf{Q1}, only two people answered correctly, while the rest four people even had no clues. In \textbf{Q2}, four out of six people provided the correct answer for ``rd\_num''; For ``wr\_num'', the situation was worse, only one person answered correctly. 
\end{itemize}

\begin{figure}[htbp]
\centering
\includegraphics[width=1\columnwidth]
{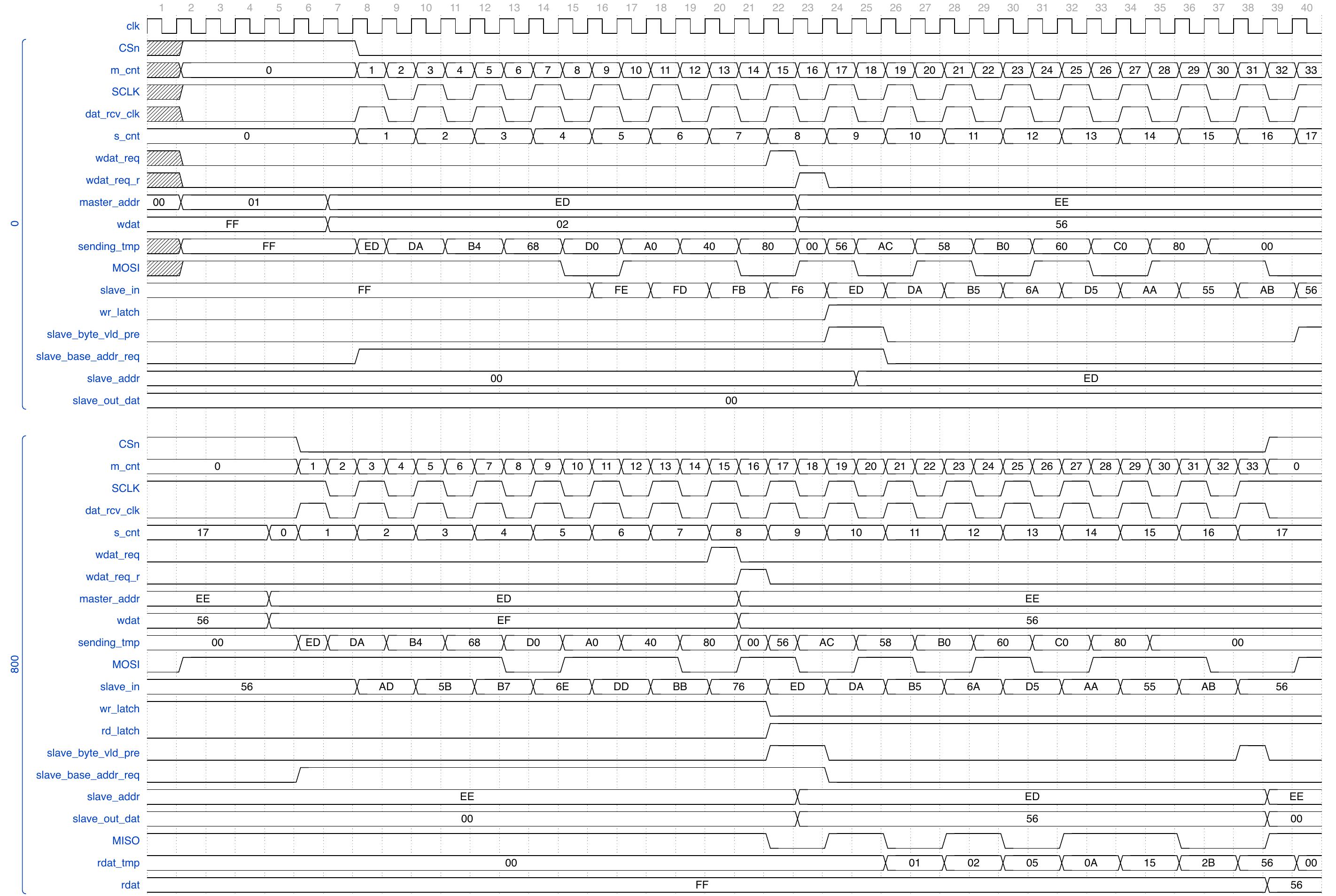}
\caption{Complete SPI protocol (slave has no internal clock)}
\label{spiProtocol}
\end{figure}

\subsubsection{SPI Protocol} The TD in Figure~\ref{spiProtocol} shows a complete SPI protocol, with a master module, a slave module, and a testbench.
\begin{itemize}
    \item \textbf{Q1 (CTA):} Please describe the transaction procedure of the protocol.
    \item \textbf{A1 (TD-Interpreter):} <<\textit{(1) Master Write}: {\color{blue}The testbench first reads data $D$ from address $X$ in the master's memory, then writes data $D$ to address $X$ in the slave's memory.} \textit{(2) Slave Feedback}: {\color{blue} The testbench takes data from address $X$ in slave's memory,} and sends back to the master to check if the written operation is correct.>>
    \item \textbf{A1 (GPT-4o):} Only describes the general transaction procedure of the SPI protocol, while overlooking the particular role of the testbench in transferring data between the master and the slave. 

    \vspace*{1mm}\item \textbf{Q2 (CTA):} How does ``sending\_tmp'' in the master get address $X$ and $D$ from the testbench? How does ``MOSI'' transmit bits sequentially?
    \item \textbf{A2 (TD-Interpreter):}  <<When ``m\_cnt'' changes from 0 to 1, ``sending\_tmp'' will read address $X$ from ``master\_addr''. {\color{blue} When ``m\_cnt'' changes from 16 to 17, by capturing the falling edge of ``wdat\_req\_r'', the master can then latch data from ``wdat'' to ``sending\_tmp''.}  ``MOSI'' transmits the ``sending\_tmp'' bit at each falling edge of ``SLCK''.>>
    \item \textbf{A2 (GPT-4o):} Wrongly captures the falling edge of ``wdat\_req'' rather than ``wdat\_req\_r'' when ``m\_cnt'' changes from 16 to 17.
    
    \vspace*{1mm}\item \textbf{Q3 (CTA):} How does the ``MISO'' signal inside the slave take the data in ``slave\_out\_dat'' from the testbench?    
    \item \textbf{A3 (TD-Interpreter):} <<This occurs during \textit{Slave Feedback}. At the falling edge of ``dat\_rcv\_clk'' while ``s\_cnt'' equals to 9, ``slave\_out\_dat'' begins to retrieve data $D$ from address $X$ in the slave’s memory. {\color{blue} Then ``MISO'' can exploit the second half duration of s\_cnt=9, to sample the highest bit of ``slave\_out\_dat'' via combinatorial logic.} Next, ``MISO'' can use the entire duration of s\_cnt=10 to sample the secondly highest bit, until the first half duration of s\_cnt=17 where "MISO" samples the lowest bit of ``slave\_out\_dat''. Since the slave has no internal clock, {\color{blue} ``MISO'' has to use combinatorial logic.}>>
    \item \textbf{A3 (GPT-4o):} The signal ``MISO'' is normally driven by sequential logic, but the SPI protocol here shows that using combinatorial logic to drive the signal ``MISO'' is more reasonable. However, GPT-4o is still insisting on using sequential logic.

    \item \textbf{Human feedback:} Four participants in our empirical study tried to analyze this TD. Although SPI is a basic peripheral protocol in embedded systems, when the participants were asked to rate the complexity of this TD, the average score reached 3.5/5 (1 means very simple, and 5 means very complex), indicating that they were confused with many details. We thus assumed most of them only used SPI in a high level, as when they had to deal with its exact transaction procedure as asked in \textbf{Q1}, they rated 3.5/5 (1 means not helpful, 5 means very helpful) for the need of interpretation. We further asked if they can identify the concrete value of $X$ and $D$ as mentioned in (\textbf{A1}), only one person answered correctly. Interestingly, only half of the participants found out the lack of internal clock in the slave module might cause problem, which means the slave module has to be driven by combinatorial logic.
\end{itemize}

\subsection{Discussion and Limitation.} From the experiments, we observed that GPT-4o faced challenges in analyzing detailed information within the TDs. For certain specific design scenarios, GPT-4o often provided general, textbook-style answers without conducting case-by-case analysis. This proves the usefulness and effectiveness of \textit{TD-Interpreter}. 

One limitation of this paper is that we adopted the \textit{wavedrom} plotting style of TDs, due to its simplicity and widespread use. Our main focus was on \textit{enabling} MLLMs to interpret TDs. Thus, developing front-end tools supporting a more diverse set of TD representations is necessary. Additionally, due to resource constraints, we were unable to construct a dataset with a larger scale. However, the most significant contribution of this work is that we provided the industry, particularly its AI-enabling departments, with a very expressive, IP-protecting, low-cost approach. In addition, using MLLMs to assist engineers in understanding TDs, may definitely appeal to many R\&D teams. Furthermore, integrating backend verification in dataset construction, and improving reasoning capability of MLLMs with reinforcement learning, are both promising directions for future research.

\section{Conclusion}
\label{sec:conclusion}

We have introduced \textit{TD-Interpreter}, a multi-modal large language model, obtained by fine-tuning LLaVA on our generated dataset of TDs and query-answer pairs. \textit{TD-Interpreter} can answer natural language questions about design and verification in timing diagrams provided as pictures. We show that \textit{TD-Interpreter} outperforms GPT-4o, and consequently the untuned LLaVA model, too, on a rich set of tasks.

\section*{Acknowledgments}
This project has received funding from the Austrian FFG and the European Key Digital Technologies Joint Undertaking (KDT JU) under grant agreement No 101097300. It is also partially supported by the Chips Joint Undertaking and the Austrian Research Promotion Agency (FFG) through the AIMS5.0 project (EU grant agreement number 101112089).


\end{document}